\documentclass[journal]{IEEEtran}
\usepackage{graphicx}
\usepackage{amsfonts}

\usepackage{hyperref}

\usepackage[english]{babel}
\usepackage{amsmath, amssymb}
\usepackage{float}
\usepackage{makecell}
\usepackage{multirow}
\usepackage{cite}
\usepackage{amsmath,amssymb,amsfonts}
\usepackage{algorithmic}
\usepackage{graphicx}
\usepackage{textcomp}
\usepackage{xcolor}

\usepackage{subfigure}
\usepackage{epstopdf}

\usepackage{graphicx}
\usepackage{comment}
\usepackage{amsmath,amssymb} 
\usepackage{color}
\usepackage{enumerate}

\usepackage{booktabs}

\ifCLASSINFOpdf
\else
\fi
\begin{document}
%
\title{Pseudo-LiDAR for Visual Odometry }
%
%
%

\author{ Huiying Deng, Guangming Wang, Zhiheng Feng,\\ Chaokang Jiang, Xinrui Wu, Yanzi Miao and Hesheng Wang
        
\thanks{*This work was supported by “the Fundamental Research Funds for the Central Universities” (2020ZDPY0303). Corresponding Author: Yanzi Miao and Hesheng Wang. The first two authors contributed equally.}
\thanks{H. Deng, C. Jiang and Y. Miao are with Advanced Robotics Research Center, AI Research Institute \& School of Information and Control Engineering, China University of Mining and Technology, Xuzhou, 221116, China. G. Wang, Z. Feng, X. Wu and H. Wang are with Department of Automation, Key Laboratory of System Control and Information Processing of Ministry of Education, Key Laboratory of Marine Intelligent Equipment and System of Ministry of Education, Shanghai Engineering Research Center of Intelligent Control and Management, Shanghai Jiao Tong University, Shanghai 200240, China.}

}

%
%

\markboth{Journal of \LaTeX\ Class Files,~Vol.~14, No.~8, August~2015}%
{Shell \MakeLowercase{\textit{et al.}}: Bare Demo of IEEEtran.cls for IEEE Journals}
%



\maketitle

\begin{abstract}
As one of the important tasks in the field of robotics and machine vision, LiDAR/visual odometry provides tremendous help for various applications such as navigation, location, etc. In the existing methods, LiDAR odometry shows superior performance, but visual odometry is still widely used for its price advantage. Conventionally, the task of visual odometry mainly rely on the input of continuous images. However, it is very complicated for the odometry network to learn the epipolar geometry information provided by the images. In this paper, the concept of pseudo-LiDAR is introduced into the odometry to solve this problem. The pseudo-LiDAR point cloud back-projects the depth map generated by the image into the 3D point cloud, which changes the way of image representation. Compared with the stereo images, the pseudo-LiDAR point cloud generated by the stereo matching network can get the explicit 3D coordinates. Since the 6 Degrees of Freedom (DoF) pose transformation occurs in 3D space, the 3D structure information provided by the pseudo-LiDAR point cloud is more direct than the image. Compared with sparse LiDAR, the pseudo-LiDAR has a denser point cloud. In order to make full use of the rich point cloud information provided by the pseudo-LiDAR, a projection-aware dense odometry pipeline is adopted. Most previous LiDAR-based algorithms sampled 8192 points from the point cloud as input to the odometry network. The projection-aware dense odometry pipeline takes all the pseudo-LiDAR point clouds generated from the images except for the error points as the input to the network. While making full use of the 3D geometric information in the images, the semantic information in the images is also used in the odometry task. The fusion of 2D-3D is achieved in an image-only based odometry. Experiments on the KITTI dataset prove the  effectiveness of our method. To the best of our knowledge, this is the first visual odometry method using pseudo-LiDAR.

\end{abstract}

\begin{IEEEkeywords}
Deep learning, pseudo-LiDAR, visual odometry.
\end{IEEEkeywords}

%
\IEEEpeerreviewmaketitle

\section{Introduction}
%
%
%
%
\begin{figure}[t]
	\centering
	\includegraphics[scale=0.56]{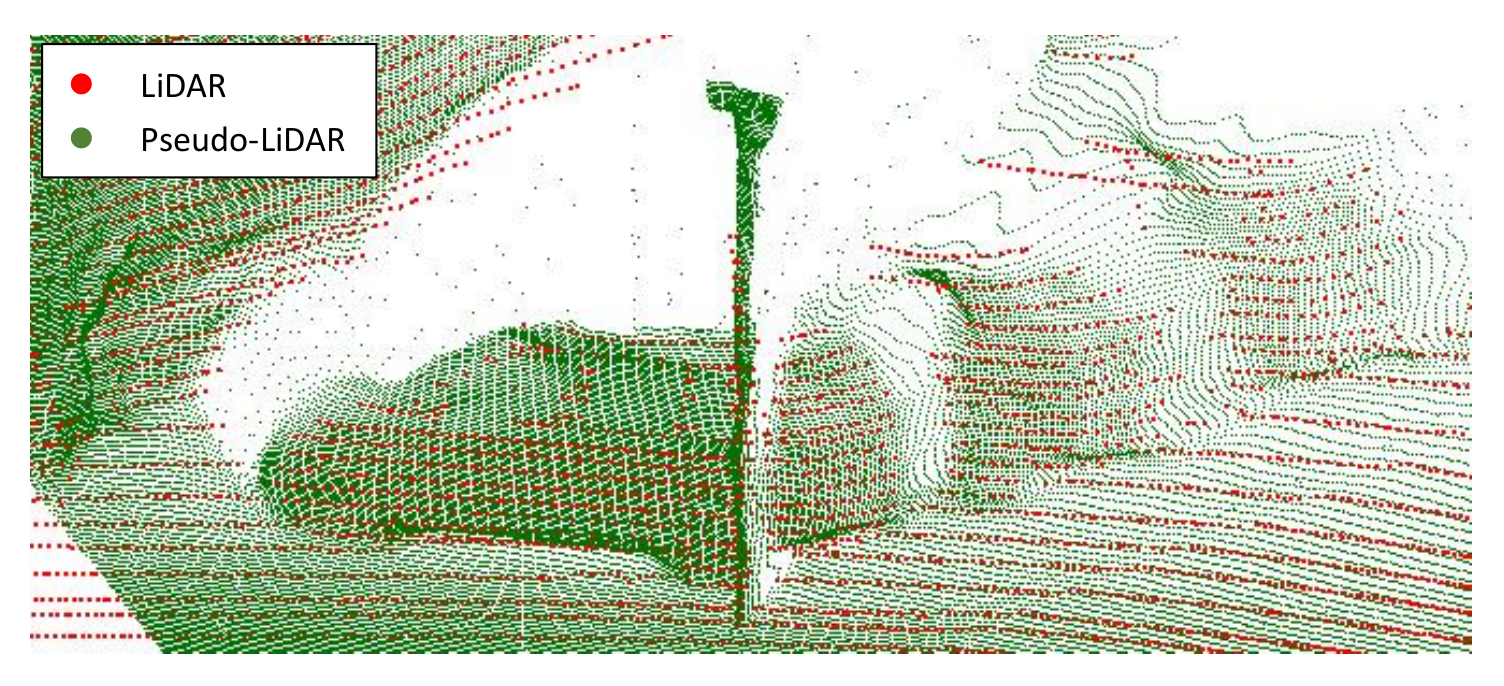}
	\vspace{-4mm}
\caption{\textbf{Comparison between the pseudo-LiDAR and LiDAR. }The image shows that the pseudo-LiDAR point cloud is denser compared to the LiDAR point cloud, and the pseudo-LiDAR and LiDAR point cloud are almost overlapping.}
	\label{fig1}
\end{figure}

\IEEEPARstart{O}{dometry} provides real-time pose information to estimate the relative position of the mobile robot. It uses an incremental method to estimate the pose of the robot. LiDAR/visual odometry receives widespread attention due to its application in autonomous driving \cite{geiger2013vision}. Traditional visual odometry works \cite{Wang_2017,Wang_2017_2} feed images directly into the odometry network. The performance of such methods is limited by the fact that the images do not provide explicit 3D information to help the odometry network learn the 6-DoF pose in space. With the development of odometry, more and more researches combine odometry with other tasks, such as depth estimation \cite{2017Unsupervised,bian2019unsupervised}. Multi-task learning is utilized to achieve a deeper understanding of the scene. The addition of depth brings rich spatial information to the odometry task. This is in line with our idea. However, unlike other methods, a different approach is taken to combining depth with odometry, i.e., pseudo-LiDAR. By learning geometry, shape, and scale information contained in 3D point clouds, the surrounding environment can be better understood. The introduction of pseudo-LiDAR allows the visual odometry to learn from explicit 3D coordinates. For the generation of pseudo-LiDAR point clouds, specifically, a stereo matching network is used to estimate the depth map and then  the generated depth map is utilized to back-project each pixel in the image into 3D space. 

Compared with the LiDAR point cloud, the pseudo-LiDAR point cloud is denser. Even with 64-beam LiDAR, the point cloud it provides is very sparse (cf. Fig. \ref{fig1}). PointNet++ \cite{qi2017pointnet++} is a good choice for learning the local and global features of the point cloud. However, the dense point cloud in the pseudo-LiDAR makes the ball query in PointNet++ \cite{qi2017pointnet++} take a lot of time. Due to the limitation of calculation power, most of the previous LiDAR-based odometry methods use 8192 points as input to the odometry network. If such a method is adopted in the pseudo-LiDAR-based odometry, a large amount of 3D geometric information in the pseudo-LiDAR is discarded. Therefore projection-aware methods are used to learn the features of dense pseudo-LiDAR point clouds efficiently. Many projection-based methods lose depth information when projecting point clouds to the 2D plane. Wang et al. \cite{wang2021efficient} assign the 3D coordinate values in the point cloud to the projected point so that the information in the point cloud is not lost while learning the point cloud efficiently. In our approach, the 3D coordinates of the pseudo-LiDAR point cloud are reprojected into their original 2D coordinates, i.e., each pixel grid is placed with its corresponding 3D coordinate values. Points at pixel locations search for their neighboring points in the fixed size kernel to aggregate features, which greatly reduces the time required.

In order to achieve a deeper understanding of the point cloud by the network, image semantic information is used to enhance the geometric features of the pseudo-LiDAR point cloud. Since there is a one-to-one correspondence between the pseudo-LiDAR and the image itself, the fusion of the image and the point cloud can be performed as long as they are both sampled in the same steps at each layer of the pyramid.

The main contributions of this paper are as follows:

\begin{itemize}
\item By introducing the concept of pseudo-LiDAR, a new framework of visual odometry based on the LiDAR algorithm is proposed. The 3D spatial information provided by the pseudo-LiDAR allows the visual odometry network to learn more directly about the 6-DoF pose in space.
\item A projection-aware odometry method is adopted to take advantage of the rich point clouds in pseudo-LiDAR.  This method achieves more efficient point cloud sampling and grouping by projecting the 3D coordinate values of the pseudo-LiDAR point cloud back into its original pixel coordinates. The ablation experiments also show that this approach improves the performance of the network.
\item To make full use of the semantic and geometric information contained in images to enhance the network's understanding of the environment, image-only based 2D-3D fusion is designed. The ablation experiments verify that the use of image semantic information to enrich point cloud features does improve the performance of odometry estimation.
\item The experimental results on KITTI dataset \cite {geiger2013vision} show that the idea of introducing projection-aware pseudo-LiDAR and image-only based 2D-3D fusion can effectively improve the performance of odometry estimation.
\end{itemize}

\begin{figure*}[t]
	\centering
	\includegraphics[scale=0.36]{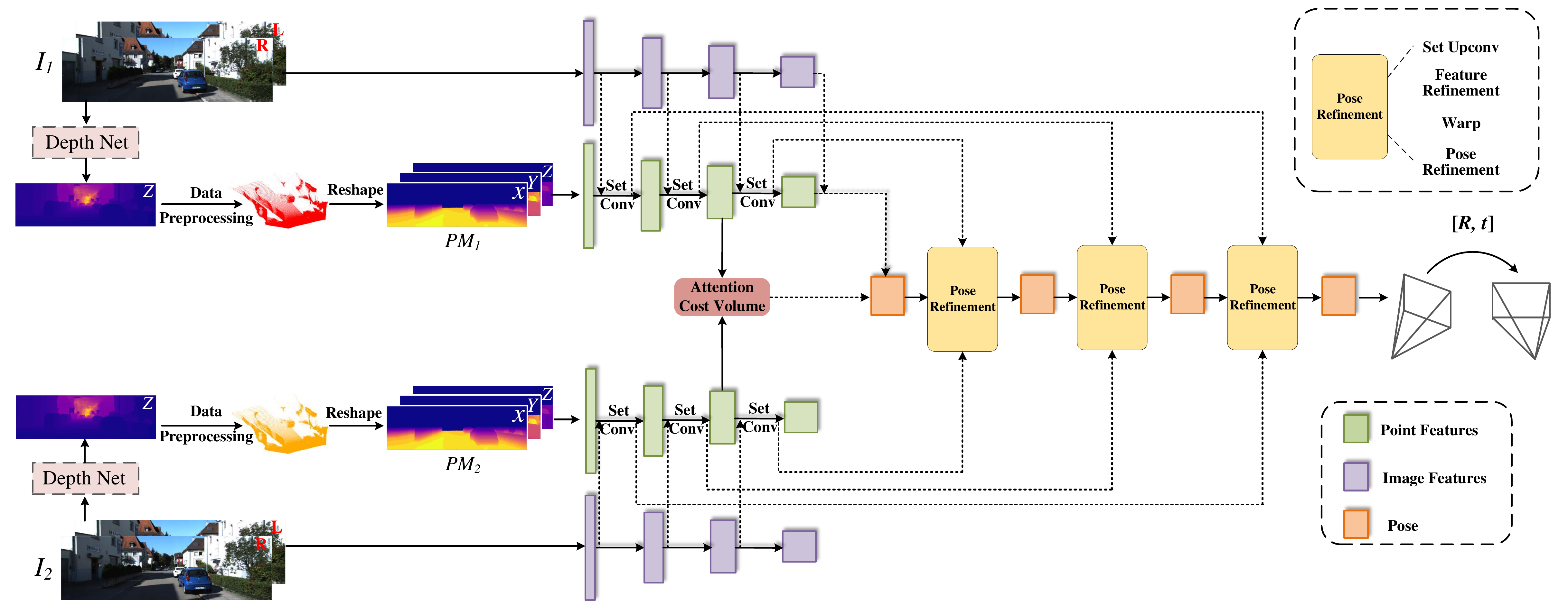}
	\vspace{-4mm}
\caption{\textbf{The architecture of our proposed odometry network.} The odometry network is divided into three parts: a depth network, a point stream and a image stream. For the input images of two consecutive frames, the depth net is used to generate the corresponding depth map. The depth map is then used to generate the pseudo-LiDAR point cloud. The pseudo-LiDAR point cloud projects its 3D coordinates ($x, y, z$) onto its own pixel coordinates ($u, v$), generating a $H\times W\times3$ point map $PM$. The generated $PM$ is used as the input of point stream, and the input of image stream is image. In the corresponding pyramid layer of point stream and image stream, the point cloud features generated by point stream are fused with the semantic features generated by image stream to enhance the point features.}
	\label{fig2}
\end{figure*}

\section{Related Work}
 
\subsection{Odometry}
Methods of visual odometry can be mainly divided into geometry-based methods \cite{Mur_Artal_2017,2011StereoScan} and deep learning-based methods \cite{Wang_2017,zou2020learning}. With the development of deep learning, Wang et al. \cite{Wang_2017} propose the first end-to-end visual odometry framework based on deep learning to achieve the camera pose. It uses convolutional neural networks (CNN) and deep recurrent neural networks (RNN) to learn image features and the relationship between sequences respectively. In subsequent works, more and more researches combine odometry with other tasks such as optical flow and depth \cite{tartanvo2020corl,2017Unsupervised,Yin_2018,bian2019unsupervised, wang2020unsupervised,wang2019unsupervised} to realize mutual promotion in multi-task joint learning. Zhou et al. \cite{2017Unsupervised} combine a depth network with a pose network. With the regressed depth and pose, the target frame can be reconstructed from the source frame, so that the photometric consistency can be used to train the networks. Similar to this method, Bian et al. \cite{bian2019unsupervised} propose to train the networks with a geometric consistency loss between the depth map of the source frame after warping and the depth map of the target frame. Unlike previous approaches, Mahjourian et al. \cite{Mahjourian_2018} use depth to convert 2D pixels into 3D point clouds. Iterative closest point (ICP) is propose to register two frames of 3D point cloud to adjust the ego-motion estimated by the network. These  methods all combine depth and pose, but the combination is only connected by loss, and the depth information is not introduced more deeply into the learning of the pose. By contrast, Tiwari et al. \cite{Tiwari_2020} integrate the depth map generated by CNN with the RGB image. SLAM is executed using the generated pseudo RGB-D features. The visual odometry network proposed by Wang et al. \cite{wang2019recurrent} use the connection of image and depth map estimated by the depth network as input to learning 6-DoF pose. Zou et al. \cite{zou2020learning} propose a pose network with a two-layer convolutional LSTM module. And the pre-extracted depth features are used as the input of the second layer of the ConvLSTM module. Li et al. \cite{li2018undeepvo} propose to use a network to estimate pose and depth simultaneously. The network is trained with stereo images to recover scale, but continuous monocular images are used for testing. Yin et al. \cite{Yin_2018} explore to combine depth, pose, and optical flow. The regressed depth and pose are used to recover rigid motion in the scene, while the non-rigid motion of dynamic objects is computed by the residual flow network.

\subsection{Pseudo-LiDAR}
Recently, the pseudo-LiDAR point cloud is widely used in the field of object detection, which improves the performance of object detection. Wang et al. \cite{wang2019pseudo} believe that the reason for the accuracy difference between image-based methods and LiDAR-based methods lies in the representation of data. Therefore, this method generates the depth map from the input RGB image and projects it into the 3D space to form the pseudo-LiDAR point cloud. Weng et al. \cite{weng2019monocular} also utilize the pseudo-LiDAR point cloud for 3D object detection. The 2D-3D bounding box consistency loss is used to deal with the noise in the pseudo-LiDAR point cloud.  To solve the point cloud quality problem in the pseudo-LiDAR point cloud interpolation, Liu et al. \cite{liu2021pseudo} propose a multi-modal feature fusion module to fuse the depth and texture information in RGB. You et al. \cite{you2019pseudo} reduce the error of pseudo-LiDAR point cloud by improving the accuracy of stereo depth estimation. In other fields, pseudo-LiDAR gradually gains attention. For example, Wang et al. \cite{wang2021unsupervised} apply pseudo-LiDAR to 3D scene flow estimation.

Inspired by these related works, pseudo-LiDAR is introduced into the odometry task. So that the visual odometry can establish the spatial connection of the points through the explicit 3D coordinates to recover the camera pose.

\section{Problem Definition}
In this section, the structure of the proposed visual odometry network will be introduced in detail. As shown in Fig. \ref{fig2}, the inputs of our network are two consecutive stereo image pairs ($I_{1R}, I_{1L}$), ($I_{2R}, I_{2L}$), and the outputs are the quaternion $q \in {\mathbb{R}^4}$ and translation vector $t \in {\mathbb{R}^3}$, which represent a transformation to the mobile camera. The pipeline proposed mainly consists of three parts: a depth network, a point stream and a image stream. The depth network is used to generate the depth map of the stereo image. The point stream and image stream generate image and point features respectively, and the two streams are connected to each other by the fusion module.

\subsection{Pseudo-LiDAR}
Given a pair of stereo images, the core problem of depth estimation is to obtain the disparity $d$ of each pixel in the reference image. To obtain the disparity, the corresponding pixels of the left and right cameras need to be matched. The disparity is the horizontal distance of the corresponding pixels in the left and right images. With the disparity, the depth $Z$ can be calculated using the camera focal length $f$ and the baselines $b$, and the formula is:
\begin{equation}
Z = \frac{f\times b}{{d}}.
\end{equation}

For the depth estimation module, in this paper, the stereo depth network GA-Net \cite{zhang2019ga} is used to estimate the disparity of the stereo images. GA-Net \cite{zhang2019ga} captures the local and the global cost dependencies through the proposed guided matching cost aggregation (GA) strategies to improve the accuracy of disparity estimation in the occlusions, large textureless and reflective regions.

There are many representations to characterize the 3D data, including depth map, point cloud, and voxel grid. Among them, the point cloud is widely used because of its simple expression and more primitive geometric information. Therefore, in this paper, instead of using the depth map directly, the pseudo-LiDAR generated from the depth map is used to obtain the 3D coordinates of each pixel in the image.

The specific conversion process is as follows. With the estimated depth and camera matrix, the 3D position ($x, y, z$) of each pixel ($u, v$) in images can be calculated. $Z$ is the estimated depth map of the pixel in the camera coordinate and ($c_U, c_V$) is the pixel location of the camera center. $f_U$ and $f_V$ are the focal length of the camera along the x and y axes, respectively. 
\begin{equation}
\left\{ \begin{array}{l}
z = Z,\\
x = \frac{{(u - {c_U}) \times z}}{{{f_U}}},\\
y = \frac{{(v - {c_V}) \times z}}{{{f_V}}}.
\end{array} \right.
\end{equation}

Points 2$m$ above the ground in the generated pseudo-LiDAR are removed as sky points with estimation errors. In addition, points at distances over 30$m$ are filtered out due to the low accuracy of depth estimation at remote points.

\begin{figure}[t]
	\centering
	\includegraphics[scale=0.68]{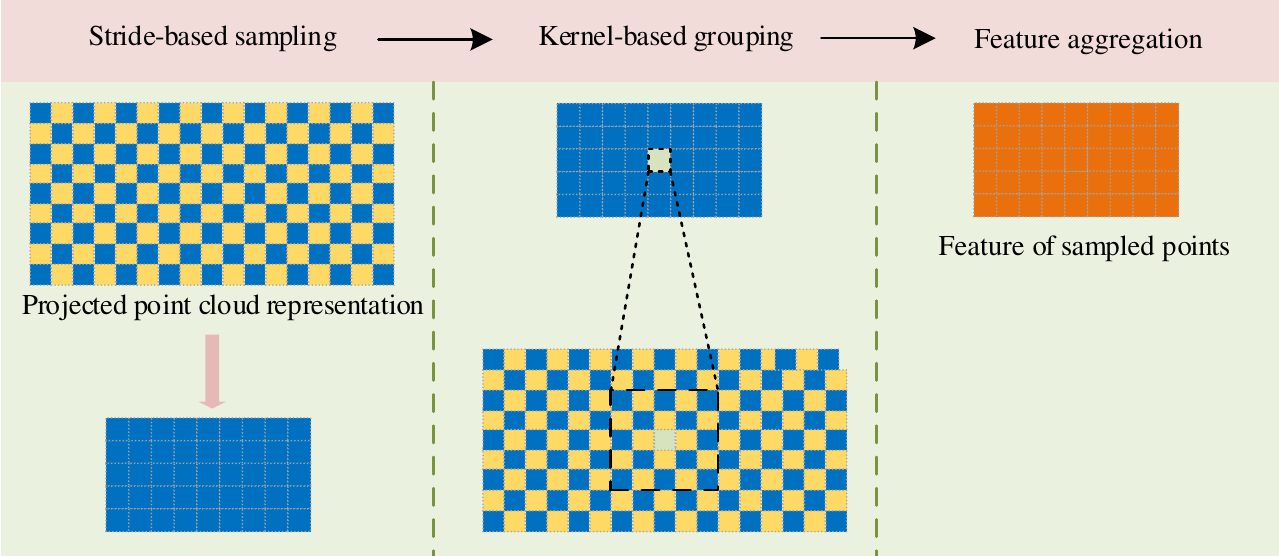}
	\vspace{-4mm}
\caption{\textbf{The details of the projection-aware set conv layer.} In order to implement downsampling and feature aggregation of the projected point cloud in the 2D plane, three steps are followed: stride-based sampling, kernel-based grouping, and feature aggregation.}
	\label{fig3}
\end{figure}

\subsection{Projection-aware Dense Odometry Net}
In the proposed network, the 3D coordinate values in the pseudo-LiDAR point cloud will be projected back to their original pixel coordinates to generate the point map $PM$ of $H \times W \times 3$ as the input of the point stream. $PM_1$ and $PM_2$ represent the point maps of two consecutive frames. The proposed network consists of the classical Pyramid, Warping, and Cost volume (PWC) structure, which allows the regressed poses to be refined in layers from coarse to fine.

{\bf Projection-aware Point Feature Pyramid: }Fig. \ref{fig3} illustrates the details of the projection-aware set conv layer. Inspired by \cite{wang2021efficient}, first, the stride-based method is used to downsample the point map $PM$. Based on the sampling step set for each pyramid layer, the point cloud at fixed intervals in the 2D plane is indexed. Then, each indexed sampled point is used as a centroid to aggregate its surrounding points. Specifically, for each centroid $x_i^c$, the corresponding point is found in the projected point cloud before sampling and a fixed size kernel is set around it. After that, K Nearest Neighbor (KNN) is used within the kernel to obtain the points $x_i^k$ ($k$ = 1, 2, ..., $K$) around the centroid. After obtaining the centroid and its surrounding grouped points, Multi-Layer Perceptron (MLP) and max pooling are used to perform feature aggregation in order to obtain the feature of the centroid. The formula is:
\begin{equation}
{f_i} = \mathop {maxpool}\limits_{k = 1,2,...K} (MLP((x_i^k - {x_i^c}) \oplus f_i^k \oplus f_i^c)),
\end{equation}
where $f_i^c$ and $f_i^K$ are the original features of $x_i^c$  and $x_i^k$ . $f_i$ is the output feature of the centroid $x_i^c$. $\oplus$ denotes the concatenation of two vectors, and MAX(·) indicates the max pooling operation.

\begin{figure}[t]
	\centering
	\includegraphics[scale=0.50]{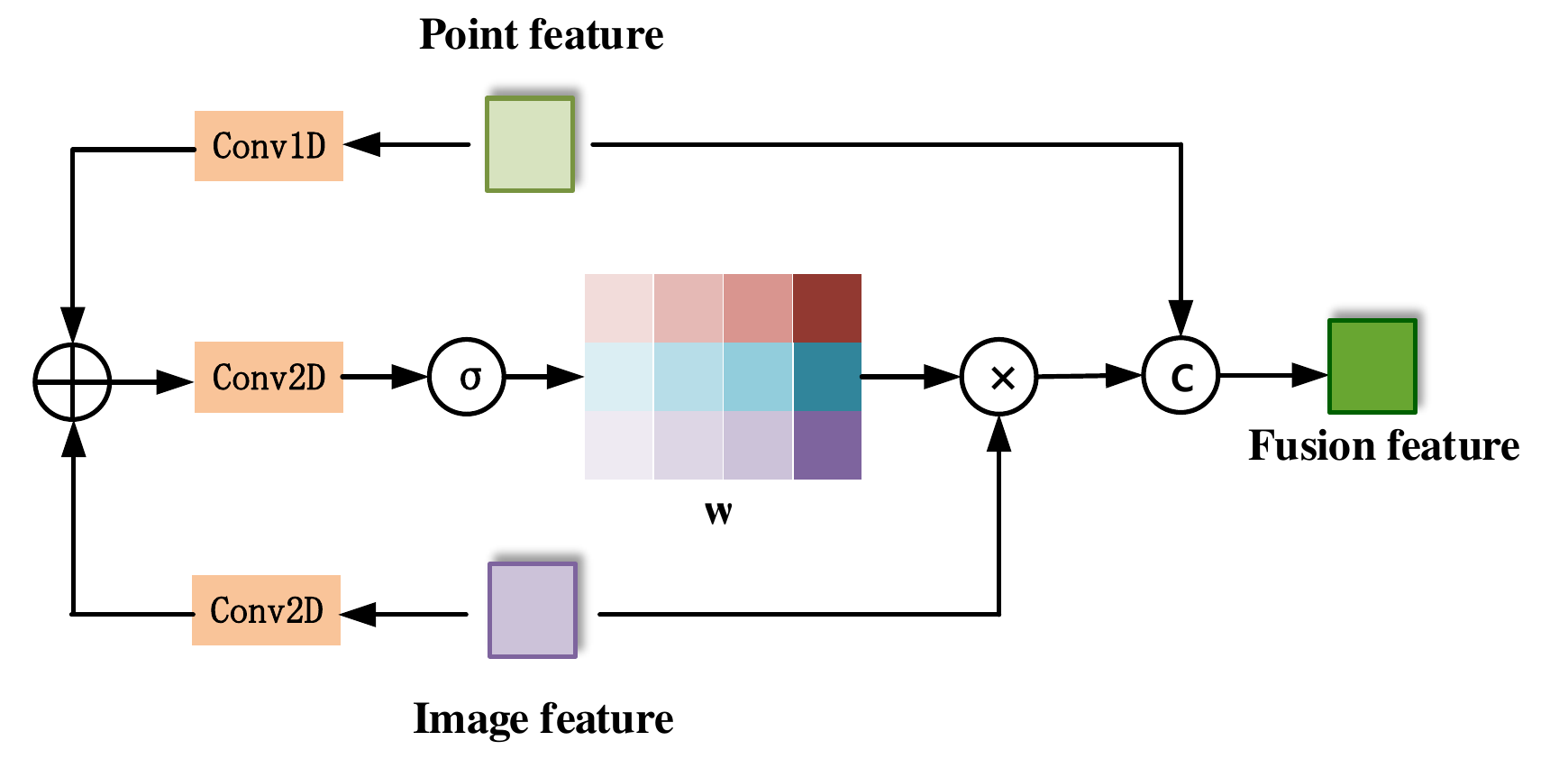}
	\vspace{-4mm}
\caption{\textbf{Structure of the module for fusion of point feature with image feature. 
	} }
	\label{fig4}
\end{figure}

{\bf 2D-3D Fusion: }
To use the semantic information in the images to enhance the network's understanding of point clouds, the point stream fuse the extracted image features with their corresponding point features. Unlike the traditional 2D-3D fusion of image and point cloud, our method does not need to use external parameters to correspond the image and point cloud, because the image and pseudo-LiDAR point cloud itself are one-to-one correspondence. At each layer of the image stream, convolution is used to extract image features, and the convolution step is set to match the sampling step of the corresponding point stream to ensure the correspondence between the image and the point map. After getting the corresponding point and image features, the image features and point features can be fused as shown in Fig. \ref{fig4}. $ \{ {{\rm{F}}^l} \in {\mathbb{R}^{{C_l} \times {H_l} \times {W_l}}}\} _{l = 1}^L$ is the features of the point map and $ \{ {{\rm{G}}^l} \in {\mathbb{R}^{{C'_l} \times {H_l} \times {W_l}}}\} _{l = 1}^L$ is the features of the image, where $l$ denotes the number of layers of these features in the pyramid, $C$ is the number of channels, and $H$ and $W$ are the height and width of the feature map, respectively. Since the semantic information provided by images is susceptible to the outdoor environment, a weight $w$ is designed, as in many other methods\cite{huang2020epnet,zhuang2021perception}, to evaluate the importance of image features on the point features. To obtain the weight $w$, 2D convolution and 1D convolution are applied to image features and point features respectively to reduce the feature channels. After that the convolved features are concatenated and then the information interaction between semantic and geometric channels is achieved using 2D convolution while performing channel reduction. Finally, the generated weights $w$ of dimension $H \times W \times 1$ are normalized using a sigmoid function. The formula is:
\begin{equation}
w = \sigma ({C_2}({C_2}({G^l}) + {C_1}({F^l}))),
\end{equation}
where $C_2(\cdot)$, $C_1(\cdot)$ represent the 2D and 1D convolution, respectively, and $\sigma(\cdot)$ is the sigmoid activation function.
Finally, the fused point cloud features can be computed by
\begin{equation}
F_{fused}^l= w \odot {G^l} + {F^l},
\end{equation}
where $\odot$ denotes the multiplication between elements.

{\bf Attentive Cost Volume: }After obtaining the point feature from the pyramid module, we use cost volume  \cite{wang2021efficient} to compute the embedding feature $E$ between $PM_1$ and $PM_2$. The embedding feature $E$ is then used to compute the mask $M$ that masks the dynamic object and regress the pose.

For $E = \{ {e_i}|{e_i} \in {\mathbb{R}^c}\} _{i = 1}^n$ and $M = \{ {m_i}|{m_i} \in {\mathbb{R}^c}\} _{i = 1}^n$, the pose can be regressed by the following calculation:
\begin{equation}
q = \frac{{FC(\sum\limits_{i = 1}^n {e_i \odot m_i} )}}{{|FC(\sum\limits_{i = 1}^n {e_i \odot m_i} )|}},
\end{equation}
\begin{equation}
t = FC(\sum\limits_{i = 1}^n {e_i \odot m_i} ),
\end{equation}
where FC is the Fully Connected layer.

{\bf Pose Refinement: }In order to achieve the hierarchical refinement of the pose in the pyramid structure, both the embedding feature and the mask need to be up-sampled from coarse to fine. Inspired by \cite{wang2021efficient}, the set upconv layer has similar three steps as the set conv layer, except that in the set upconv layer we use dense points as centroids to aggregate sparse points. With the set upconv layer, we can get the $l$-layer coarse embedding feature $CE^l$ and coarse mark $CM^l$. The re-embedding feature $RE^l$ is obtained by calculating the cost volume between the point cloud after the warp and the original point cloud. To calculate the projected point cloud after warp, taking the $l$-th layer as an example, two frames of the projected point cloud  $PM_1$, $PM_2$ are converted to the 3D point cloud form $PC_1$, $PC_2$. First, $PC_1^l$ is warped by the pose $T^{l+1}$ estimated from the ($l+1$)-th layer to get $PC_{1,warp}^l$.
\begin{equation}
PC_{1,warp}^l= T^{l+1}PC_1^l.
\end{equation}
For the generated $PC_{1,warp}^l$, to get the point cloud in the projected form, it is necessary to reproject it and obtain the new coordinates of each point cloud in $PC_{1,warp}^l$ on the 2D plane. Unlike \cite{wang2021efficient}, which uses cylindrical projection to get the point cloud projection after warp, we can project the point cloud directly to the 2D image plane. It is worth noting that not all point clouds will be reprojected onto the image plane due to errors in the regressed pose. Those points that are projected outside the image plane will be discarded. Using coarse embedding feature $CE^l$,corase embedding mask $CM^l$, re-embedding feature $RE^l$ , and point feature $F_{fused}^l$, $l$-layer embedding feature $E^l$ and embedding mask $M^l$ can be obtained by MLP.
\begin{equation}
e_i^l = MLP(ce_i^l \oplus re_i^l \oplus f_{fused{_i}}^l),
\end{equation}
\begin{equation}
{M^l} = soft\max (MLP({E^l} \oplus C{M^l} \oplus F_{fused}^l)).
\end{equation}

Finally, using the embedding feature and the embedding mask of the $l$-layer, the residual pose $q_{res}$ and $t_{res}$ can be calculated by Equations (6), (7). The pose of the $l$-layer is:

\begin{equation}
{q^l} = q_{res}^l{q^{l + 1}},
\end{equation}
\begin{equation}
[0,{t^l}] = q_{res}^l[0,{t^{l + 1}}]{(q_{res}^l)^{ - 1}} + [0,t_{res}^l].
\end{equation}

\subsection{Loss}
In this paper, the ground truth translation $t_{g_t}$ and quaternion $q_{g_t}$ of the camera pose is used as a supervisory signal to train the odometry network. As in \cite{wang2021pwclo}, the translation $t^l$ and quaternion $q^l$ output by the network should occupy different positions in the loss function due to their different scales and units. For this reason, two learnable parameters $s_x$ and $s_q$ are introduced. Here the loss function of the $l$-th layer is:
\begin{equation}
\begin{aligned}
{\ell^l} = \left\| {{t_{gt}} - {t^l}} \right\|\exp ( - {s_x}) + {s_x}+ \\ \qquad{\left\| {{q_{gt}} - \frac{{{q^l}}}{{\left\| {{q^l}} \right\|}}} \right\|_2}\exp ( - {s_q}) + {s_q},
\end{aligned}
\end{equation}
where $\left\| \cdot \right\|$ and $\left\| \cdot \right\|_2$ represent $\ell_1$-norm and $\ell_2$-norm, respectively. Then, for the four layers of the pyramid. The total training loss is:
\begin{equation}
\ell = \sum\limits_{l = 1}^L {{a^l}{\ell^l}}. 
\end{equation}

\setlength\tabcolsep{0.7mm}
\begin{table*}[ht]
	\begin{center}
		\caption{Result on KITTI dataset. Use sequence 00-08 as the training set and sequence 09, 10 as the testing set. $t_{rel}$ denotes translation error ($\%$) and $r_{rel}$ denotes rotation error ($deg/m$). The best results are shown in bold}
		\label{table:flyingthing}
		\begin{tabular}{clcccccccc}
			\toprule
			     &\small{Method}
			     &\multicolumn{2}{c}{\small{Seq.09}} &\multicolumn{2}{c}{\small{Seq.10}} &\multicolumn{2}{c}{\small{Mean}} \\
			     &     & 
			     $t_{rel}$& $r_{rel}$
			     & $t_{rel}$& $r_{rel}$
			     & $t_{rel}$& $r_{rel}$\\
			     \cline{1-9}\noalign{\smallskip}
			     \multirow{-0.5}{*}{\begin{tabular}[c]{@{}c@{}}
			     \small{Geometry-based VO}  \end{tabular}}
			& \small{ORB-SLAM2 (w/ LC) \cite{Mur_Artal_2017}}
			& \small{2.88}&\small{0.0025}
			& \small{3.30}& \bf\small{0.0030}
			& \small{3.09}& \bf\small{0.0028}\\
			& \small{ORB-SLAM2 (w/o LC) \cite{Mur_Artal_2017}}
			& \small{9.30}& \small{0.0026} 
			& \small{2.57}& \small{0.0032} 
			& \small{5.93}& \small{\bf0.0028} \\
			\midrule
			& \small{vid2depth \cite{Mahjourian_2018}}
			& \small{---} &  \small{---}
			& \small{21.54}&  \small{0.125}
			&  \small{21.54}&\small{0.125}
			\\\multirow{4.5}{*}{\begin{tabular}[c]{@{}c@{}}\small{Learning + Geometry VO}  \end{tabular}}
			& \small{DF-VO \cite{zhan2020visual}}
			& \small{2.07}& \bf\small{0.0023} 
			& \small{2.06}& \small{0.0036} 
			& \small{2.07}& \small{0.0030}\\
			& \small{Zhu et al. \cite{zhu2018robustness}} 
			& \small{4.66} & \small{0.017}
			& \small{6.30}& \small{0.016} 
			&\small{5.48} &\small{ 0.016}\\
			& \small{DeepMatchVO \cite{Shen_2019}}
			& \small{9.91} & \small{0.038}                       & \small{12.18}& \small{0.059}
			& \small{11.05}& \small{0.049}\\
			& \small{Gordon et al. \cite{gordon2019depth}}
			& \small{3.10}& \small{---}
			& \small{5.40} & \small{---} 
			& \small{4.25}& \small{---}\\
			& \small{pRGBD-Refined \cite{Tiwari_2020}}
			& \small{4.20}& \small{0.01}
			&\small{4.40}& \small{0.016}
			&\small{4.30} & \small{0.013}\\
			\midrule\midrule
			& \small{SfMLearner \cite{2017Unsupervised}}
			& \small{8.28}& \small{0.031}
			& \small{12.20}& \small{0.030}
			& \small{10.24}& \small{0.031}\\
			& \small{GeoNet \cite{Yin_2018}}
			& \small{28.72}&  \small{0.098}
			&\small{23.90}& \small{0.090} 
			&\small{26.31}& \small{0.094}\\
			&\small{UnDeepVO \cite{li2018undeepvo}}
			& \small{7.01}& \small{0.036}
			& \small{10.63}&\small{0.047 }
			&\small{8.82}&\small{0.042 } \\
			&\small{depth-vo-feat \cite{Zhan_2018}}
			&\small{ 11.92}& \small{0.036}
			&\small{12.62}&\small{ 0.034}
			&\small{12.27}&\small{0.035}\\
			&\small{SAVO \cite{Li_2019_2}}
			&\small{9.52}& \small{0.036}
			&\small{6.45}& \small{0.024}
			&\small{7.99}& \small{0.030}\\
			&\small{Monodepth2-M \cite{Godard_2019}}
			&\small{11.47}& \small{0.032}
			&\small{7.73}& \small{0.034}
			&\small{9.6}&\small{0.033}\\
			&\small{SC-SfMLearner \cite{bian2019unsupervised}}
			&\small{11.2}&\small{0.034}
			&\small{10.1}&\small{0.050}
			&\small{10.65}&\small{0.042}\\
			\multirow{-4}{*}{\begin{tabular}[c]{@{}c@{}}\small{self-Supervised}  \end{tabular}} 
			&\small{Wang et al. \cite{wang2019recurrent}}
			&\small{9.88}&\small{0.034} 
			&\small{12.24}&\small{0.052} 
			&\small{11.06} &\small{0.043}\\
			&\small{Li et al. \cite{li2019pose}}
			&\small{5.89}& \small{0.033}
			&\small{4.79}&\small{0.0083}
			&\small{5.34}& \small{0.021}\\
			&\small{CC \cite{Ranjan_2019}}
			&\small{6.92}&\small{0.018}
			&\small{7.97}&  \small{0.031}
			&\small{7.45}&\small{0.025}\\
			&\small{Zou et al. \cite{zou2020learning}}
			&\small{3.49}&\small{0.010}
			&\small{5.81}&\small{0.018}
			&\small{4.65}&\small{0.014} \\
			&\small{CM-VO\cite{2021Unsupervised}}
			&\small{9.69}& \small{0.034}
			&\small{10.01}& \small{0.049}
			& \small{9.85}& \small{0.0415}\\
			\midrule
			\multirow{2}{*}{\begin{tabular}[c]{@{}c@{}}\small{Supervised}  \end{tabular}}
			& \small{DeepV2D \cite{teed2018deepv2d}}
			&\small{8.71}&\small{0.037}
			&\small{12.81}&\small{0.083}
			&\small{10.76}&\small{0.037} & \\
			& \small{Ours} 
			&\bf\small{0.98}&\small{0.0043}
			&\bf\small{1.66}&\small{0.0079}
			&\bf\small{1.32}&\small{0.0061} \\
			 \bottomrule
		\end{tabular}
	\end{center}
\end{table*}

\setlength{\tabcolsep}{0.7mm}
\begin{table*}[t]
	\begin{center}
		\caption{Result on KITTI dataset. Use sequence 00, 02, 08, 09 as the training set and sequence 03, 04, 05, 06, 07, 10 as the testing set. $t_{rel}$ denotes translation error ($\%$) and $r_{rel}$ denotes rotation error ($deg/m$). The best results are shown in bold}
		\label{table:flyingthing3d}
		\begin{tabular}{clccccccccccccccc}
			\toprule
			       & \small{Method}  & \multicolumn{2}{c}{\small{Seq.03}}& \multicolumn{2}{c}{\small{Seq.04}}& \multicolumn{2}{c}{\small{Seq.05}}& \multicolumn{2}{c}{\small{Seq.06}}& \multicolumn{2}{c}{\small{Seq.07}} & \multicolumn{2}{c}{\small{Seq.10}} &  \multicolumn{2}{c}{\small{Mean}} \\     &     & $t_{rel}$   & $r_{rel}$                        & $t_{rel}$               & $r_{rel}$                          & $t_{rel}$               & $r_{rel}$                        & $t_{rel}$               & $r_{rel}$& $t_{rel}$               & $r_{rel}$  & $t_{rel}$& $r_{rel}$ & $t_{rel}$& $r_{rel}$                             \\ \cline{1-16}\noalign{\smallskip}

			& \small{DeepVO \cite{Wang_2017}}
			& \small{8.49} & \small{0.069}
			& \small{7.19}& \small{0.070}
			& \small{2.62} & \small{0.036}
			& \small{5.42}& \small{0.058}
			& \small{3.91} & \small{0.046}
			& \small{8.11}& \small{0.088}
			& \small{5.96}& \small{0.060} \\
			&\small{ESP-VO \cite{Wang_2017_2}}
			&\small{6.72} & \small{0.065}
			& \small{6.33}& \small{0.061}
			& \small{3.35} & \small{0.049}
			& \small{7.24}& \small{0.073}
			& \small{3.52} & \small{0.050} 
			& \small{9.77}& \small{0.102}
			& \small{6.16}& \small{0.066}\\
			& \small{GFS-VO \cite{2019Guided}}
			& \small{5.44} & \small{0.033 }
			& \small{2.91}& \small{0.013}
			& \small{3.27} & \small{0.016}
			& \small{8.50}& \small{0.027}
			& \small{3.37} & \small{0.023}
			& \small{6.32}& \small{0.023 }
			& \small{4.97 } & \small{0.023 }\\
			& \small{GFS-VO-RNN \cite{2019Guided}}
			& \small{6.36} & \small{0.036}
			& \small{5.95}& \small{0.024}
			& \small{5.85} & \small{0.026}
			& \small{14.58}& \small{0.050}
			& \small{5.88} & \small{0.026}
			& \small{7.44}& \small{0.032} 
			& \small{7.68}& \small{0.032}\\
			&\small{BRNN \cite{2019Visual}}
			& \small{4.74} & \small{0.031}
			& \small{3.90 }& \small{0.022 }
			& \small{6.02} & \small{0.027}
			& \small{9.59}& \small{0.024}
			& \small{5.28}& \small{0.023}
			& \small{7.04}& \small{0.033} 
			& \small{6.10} & \small{0.026} \\
			\multirow{-5}{*}{\begin{tabular}[c]{@{}c@{}}\small{Supervised}  \end{tabular}} 
			& \small{BeyondTracking \cite{xue2019beyond}}
			& \small{3.32} & \small{0.021}
			&\small{ 2.96}& \small{0.018}
			& \small{2.59} & \small{0.013}                       & \small{4.92}& \small{0.019}
			& \small{3.07}& \small{0.018}
			& \small{3.94}& \small{0.017}
			& \small{3.47}& \small{0.018}\\
			& \small{Ours}
			& \small{\bf1.85}& \small{\bf0.0085}
			& \small{\bf1.40}& \small{\bf0.011}
			& \bf\small{1.79} & \bf\small{0.011}
			& \bf\small{1.67}& \bf\small{0.0092}
			& \bf\small{2.55}& \bf\small{0.015}
			& \bf\small{1.87}  & \bf\small{0.0092}
			& \small{\bf1.86}& \small{\bf0.011}\\
			 \bottomrule
		\end{tabular}
	\end{center}
\end{table*}

\section{Experiments}
\subsection{Dataset and Metrics}
The proposed odometry network is trained and evaluated on the KITTI odometry dataset \cite {geiger2013vision}. The KITTI odometry dataset   \cite {geiger2013vision} contains 22 independent stereo sequences. Among them, 11 sequences (00-10) contain the ground truth (trajectory), and the remaining 11 sequences (11-21) do not provide ground truth. Like many other methods, the sequence 00-10 is used for training and evaluation. For the data set segmentation method, to facilitate the comparison with other methods, two common methods have been adopted. One is to use the sequence 00-08/09, 10 for training/testing respectively, as shown in Table~\ref{table:flyingthing}. The other is to use the long sequence 00, 02, 08, 09 for training, and the short sequence 03, 04, 05, 06, 07, 10 for testing, as shown in Table~\ref{table:flyingthing3d}.

To quantitatively show the effect of our method and make it easier to compare with other methods, the evaluation standard proposed by the KITTI odometry benchmark is used. For each test sequence, the average translational and rotational errors on all possible subsequences in the length of 100, 200, ..., 800m are evaluated, and errors are measured in percent (for translation) and degrees per meter (for rotation).

\subsection{Network Details}
Our model is trained supervised on the KITTI dataset \cite {geiger2013vision}. For depth network, the pre-trained model provided by GA-Net \cite{zhang2019ga} is used. During the training process, the depth network will not be trained. The pseudo-LiDAR point cloud generated by the depth map will be saved to reduce unnecessary training time. An Quadro RTX 3090 is used to train our complete model. The Adam optimizer is adopted with $\beta_1 = 0.9$, $\beta_2 = 0.999$. The initial learning rate is set to 0.001, which decays to 0.00001 during training. Our decay step is set to attenuate the learning rate by 0.7 every 13 epochs. Similar to PWCLO-Net \cite{wang2021pwclo}, the initial values of the trainable parameters $s_x$ and $s_q$ are set to 0.0 and -2.5, respectively. For Equation (14), $a_1=1.6$, $a_2=0.8$, $a_3=0.4$, and $L=4$. The batchsize size is 8.

\begin{figure}[t]
	\centering
	\includegraphics[scale=0.35]{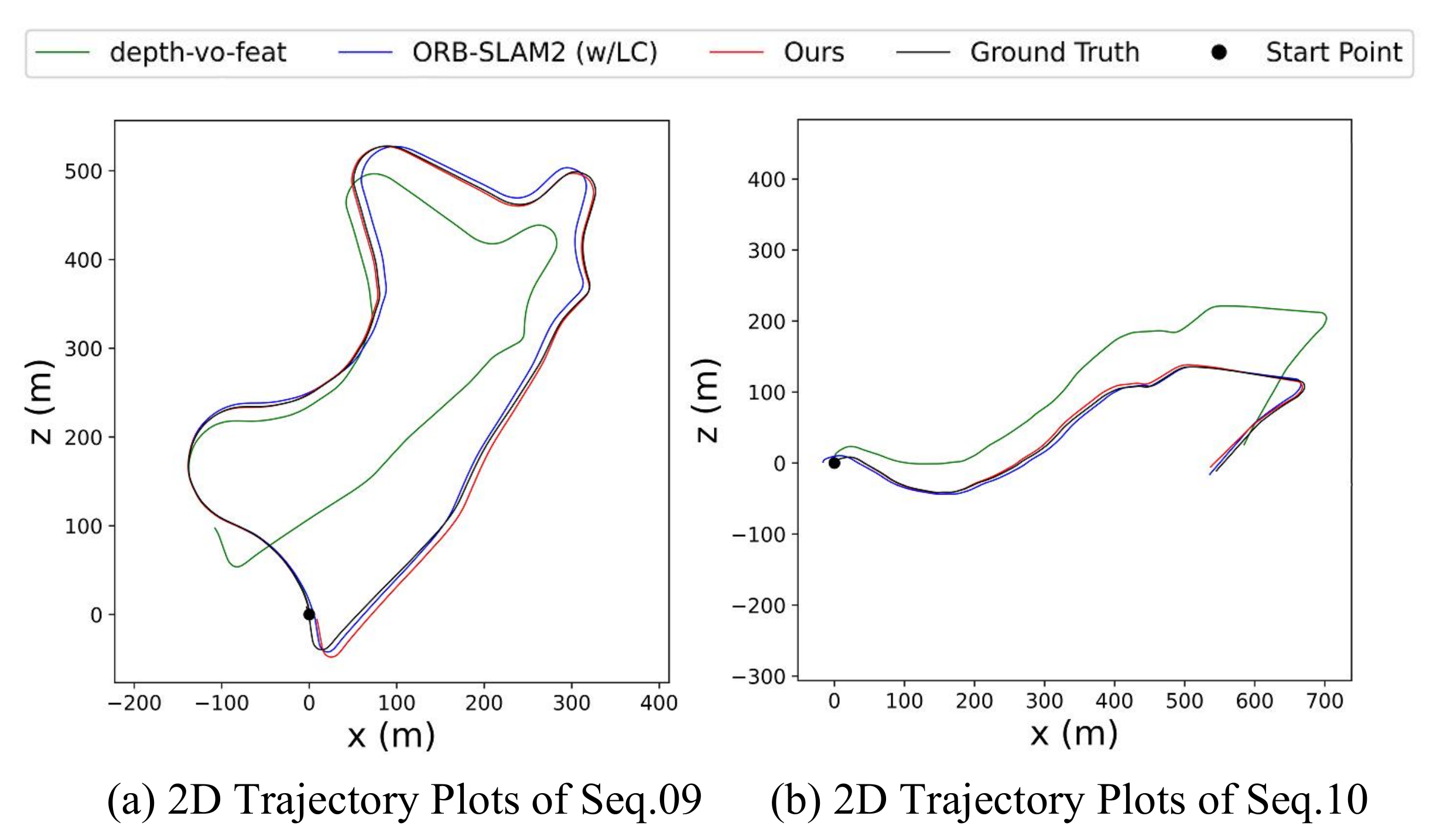}
	\vspace{-4mm}
\caption{\textbf{The 2D trajectory results on KITTI sequence 09 and 10 with ground truth. 
	} Trajectory results of depth-vo-feat,  ORB-SLAM2 (w/ LC) and ours on KITTI. }
	\label{fig:seq1}
\end{figure}

\begin{figure}[t]
	\centering
	\includegraphics[scale=0.34]{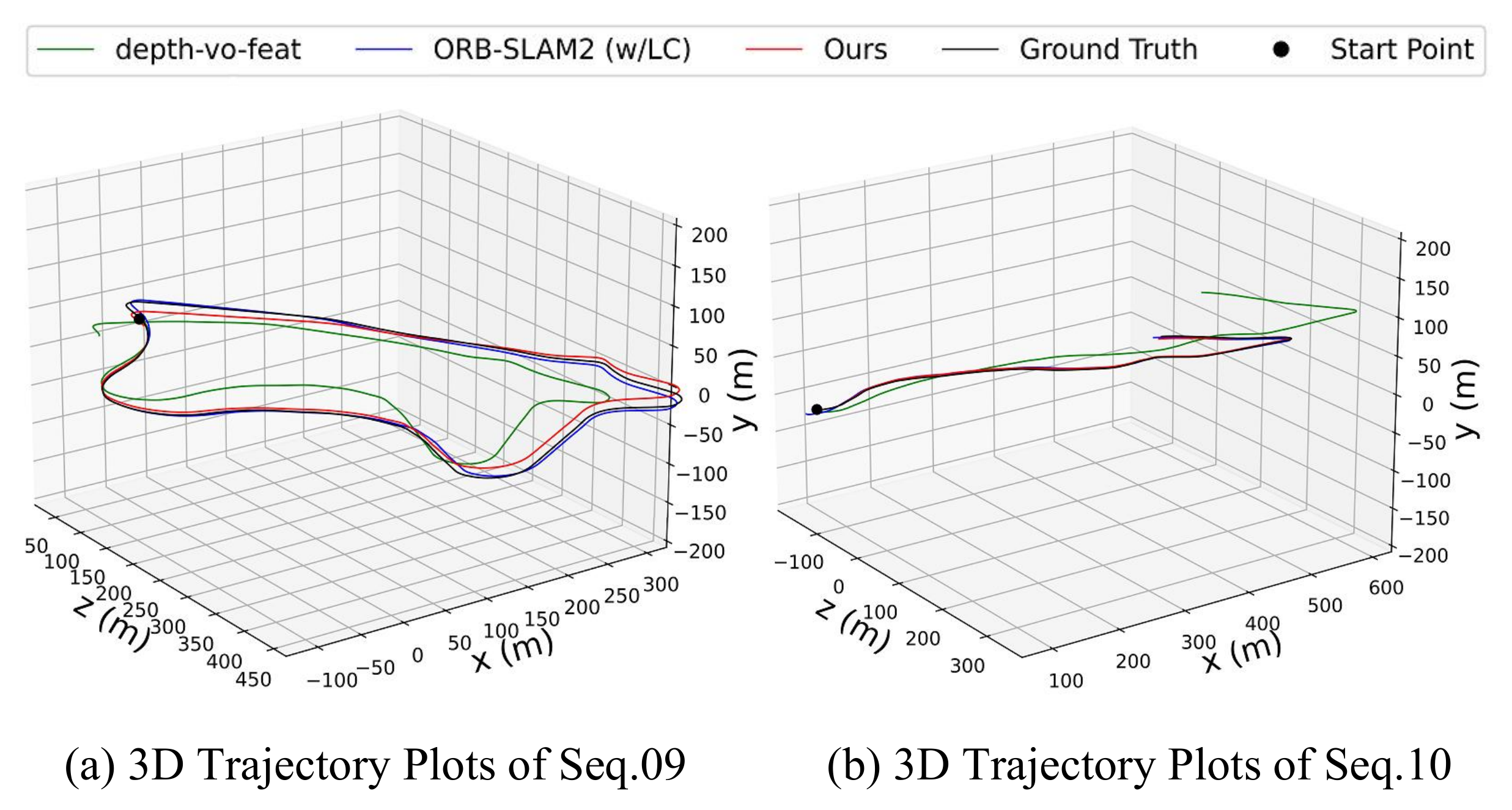}
	\vspace{-4mm}
\caption{\textbf{The 3D trajectory results on KITTI sequence 09 and 10 with ground truth. 
	} Trajectory results of depth-vo-feat,  ORB-SLAM2 (w/ LC) and ours on KITTI. }
	\label{fig:seq2}
\end{figure}

\begin{figure}[t]
	\centering
	\includegraphics[scale=0.58]{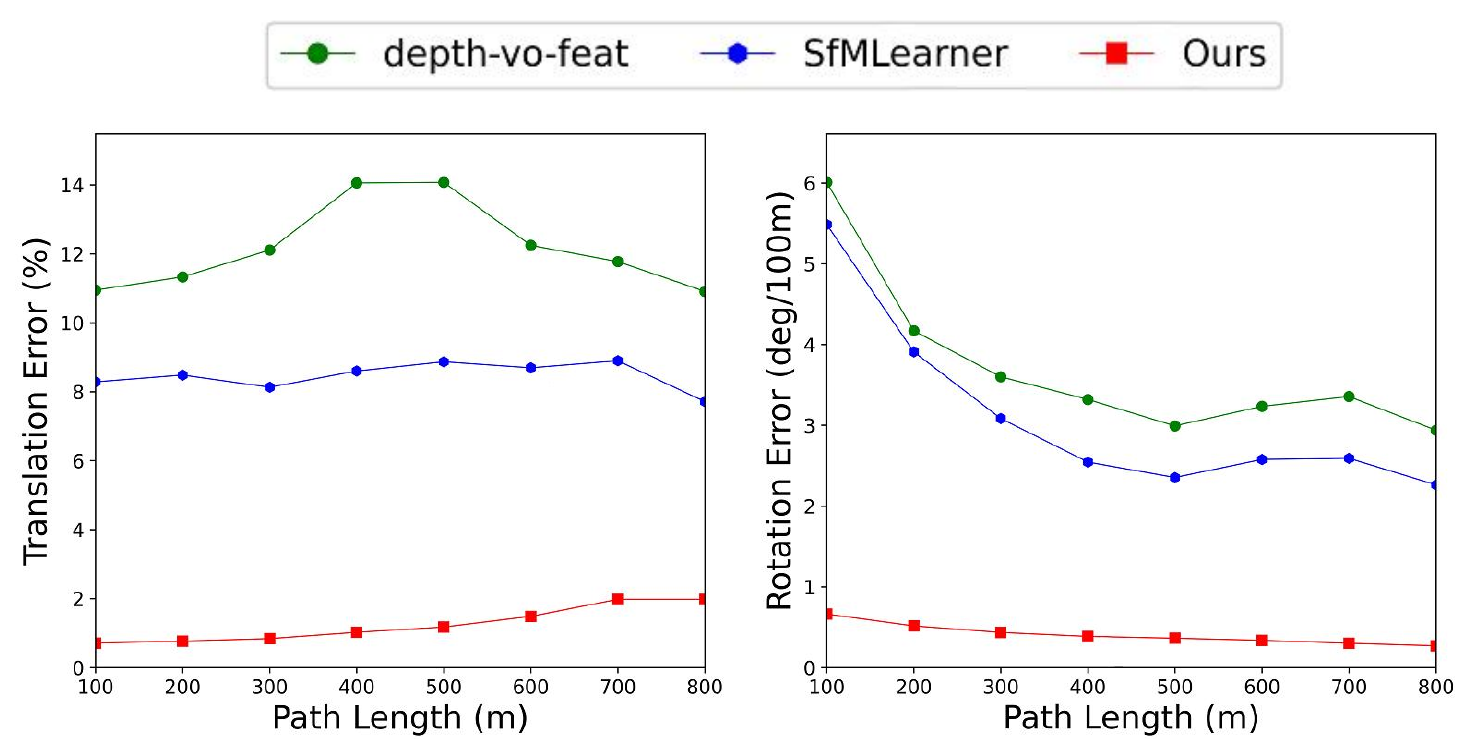}
	\vspace{-4mm}
\caption{\textbf{Translational and rotational error on KITTI sequences. 
	} Average translational and rotational error on KITTI sequences 09 on  all possible subsequences in the length of 100, 200, ..., 800m.}
	\label{fig:seq3}
\end{figure}

\section{Experimental Results}
In this section, the quantitative and qualitative results of the visual odometry network performance are shown in tables and pose visualizations.

\subsection{Performance Evaluation}
In the existing related methods, there are mainly two modes to divide the training/testing set. In order to make a fair comparison with the related methods, our model is trained/tested in these two modes respectively.

\textbf{Using sequences 00-08/09-10 as training/testing sets:}
Quantitative results are summarized in Table~\ref{table:flyingthing}. ORB-SLAM2 \cite{Mur_Artal_2017}  with loop closing (w/ LC) / without loop closing (w/o LC) is a classical feature point based geometric method. vid2depth \cite{Mahjourian_2018}, DeepMatchVO \cite{Shen_2019}, SfMLearner \cite{2017Unsupervised}, GeoNet \cite{Yin_2018}, UnDeepVO \cite{li2018undeepvo}, depth-vo-feat \cite{Zhan_2018}, Monodepth2-M \cite{Godard_2019}, SC-SfMLearner \cite{bian2019unsupervised} and CC \cite{Ranjan_2019} are all combined depth estimation with odometry, but these methods only use the loss to connect the deep network with the pose network. In the methods Gordon et al. \cite{gordon2019depth}, pRGBD-Refined \cite{Tiwari_2020}, SAVO \cite{Li_2019_2}, Wang et al. \cite{wang2019recurrent}, Zhu et al.\cite{zhu2018robustness}, Li et al. \cite{li2019pose}, 
and DeepV2D \cite{teed2018deepv2d}, depth information is used in the pose network in different ways to optimize the pose. But unlike our approach where depth is directly used to convert a 2D image to a 3D point cloud, pRGBD-Refined \cite{Tiwari_2020} concatenates the regressed depth map with the image into pseudo RGB-D to estimate the pose. While these methods all exploit depth information in images,our method acts directly on the 3D coordinates, where scale information between points is revealed. So points that are far apart in real 3D space are not put together for feature aggregation. Previous methods do not exploit the explicit 3D structure, and depth was used in an indirect form. The comparison results also show the superiority of our method.

\textbf{Using sequences 00,02,08,09/03,04,05,06,07,10 as training/testing sets:}
Quantitative results are summarized in Table~\ref{table:flyingthing3d}. DeepVO \cite{Wang_2017} propose a visual odometry framework based on deep learning, which realize the end-to-end estimation of the camera pose. On the basis of DeepVO \cite{Wang_2017}, FC layer and SE(3) composition layer are added to ESP-VO \cite{Wang_2017_2} to directly estimate a range of poses and uncertainties. Both DeepVO \cite{Wang_2017} and ESP-VO \cite{Wang_2017_2} do not provide explicit depth information to the odometry network, and the pose network needs to learn indirectly from images, which is much more complicated than directly learning the input 3D information to estimate pose. 

The qualitative results are shown in Fig. \ref{fig:seq1}, \ref{fig:seq2} and \ref{fig:seq3}. we compare with the two methods ORB-SLAM2 (w/ LC) \cite{Mur_Artal_2017} and depth-vo-feat \cite{Zhan_2018}. The average rotation and translation errors for different path lengths are shown in Fig. \ref{fig:seq3}. As the path increases, the accumulated error of the odometry is effectively controlled, and our method also shows robustness to the error.

\setlength{\tabcolsep}{0.7mm}
\begin{table}[t]
	\begin{center}
		\caption{Ablation Study on the KITTI dataset. P denotes that the input to the point stream is 8192 randomly sampled points from the pseudo-LiDAR point cloud. PA indicates the projection-aware approach, and 2D-3D indicates image-only based 2D-3D fusion.}
		\label{table3}
		\begin{tabular}{clcc|ccccccccccccc}
			\toprule
			       & \small{P}  &\small{PA}
			       &\small{2D-3D} & \multicolumn{2}{c}{\small{Seq.09}} & \multicolumn{2}{c}{\small{Seq.10}} &  \multicolumn{2}{c}{\small{Mean}} \\     
			       &     &     &     &
			       $t_{rel}$   & $r_{rel}$& 
			       $t_{rel}$  & $r_{rel}$&
			       $t_{rel}$ & $r_{rel}$                          \\ \cline{1-10}\noalign{\smallskip}

			& \small{\checkmark}& \small{} & \small{}
			& \small{2.38}& \small{0.0091}
			& \small{2.94} & \small{0.018}
			& \small{2.66}& \small{0.014}\\
			&\small{}& \small{\checkmark} & \small{}
			& \small{2.01}& \small{0.0091}
			& \small{1.99} & \small{0.013}
			& \small{2.00}& \small{0.011}\\
			& \small{}& \small{\checkmark} & \small{\checkmark}
			& \bf\small{0.98}& \bf\small{0.0043}
			& \bf\small{1.66} & \bf\small{0.0079}
			& \bf\small{1.32}& \bf\small{0.0061}\\
			 \bottomrule
		\end{tabular}
	\end{center}
\end{table}

\subsection{Ablation Study}
To verify the effectiveness of our proposed method, ablation experiments in Table~\ref{table3} are performed in the same experimental setting. The projection-aware approach has a better performance compared to the network that uses 8192 points as input. This indicates that the projection-aware approach can indeed learn the odometry more effectively for denser point cloud. And the projection-aware network with the addition of image semantic information shows better performance. This shows that richer features can provide the network with a more comprehensive scene understanding. As shown in Table~\ref{table3}, all our proposed methods improve the performance of the network.

\section{Discussion and Conclusion}
In this paper, a projection-aware pseudo-LiDAR-based visual odometry network is presented. Considering that conventional visual odometry does not make full use of the geometric structure information in images, pseudo-LiDAR is introduced. On this basis, the projection-aware approach is proposed to efficiently utilize the rich point cloud information in pseudo-LiDAR. Also, the semantic features of the images are utilized to enhance the features of the pseudo-LiDAR point cloud to help the odometry network better understand the environment. The results on KITTI dataset, as well as ablation study, show the effectiveness of our method.


\bibliographystyle{IEEEtran}
\bibliography{IEEEabrv,bare_jrnl}
\begin{IEEEbiography}[{\includegraphics[width=0.9in,height=1.3in,clip]{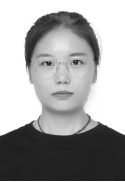}}]{Huiying Deng}
received the B.S. degree from the Department of Automation, Nanjing University of Aeronautics and Astronautics Jincheng College, Nanjing, China, in 2018. She is currently pursuing the M.E. degree in  control engineering with China University of Mining and Technology. Her current research interests include visual odometry and 3D LiDAR odometry.
\end{IEEEbiography}
 \vspace{-10mm}
\begin{IEEEbiography}[{\includegraphics[width=1in,height=1.25in,clip,keepaspectratio]{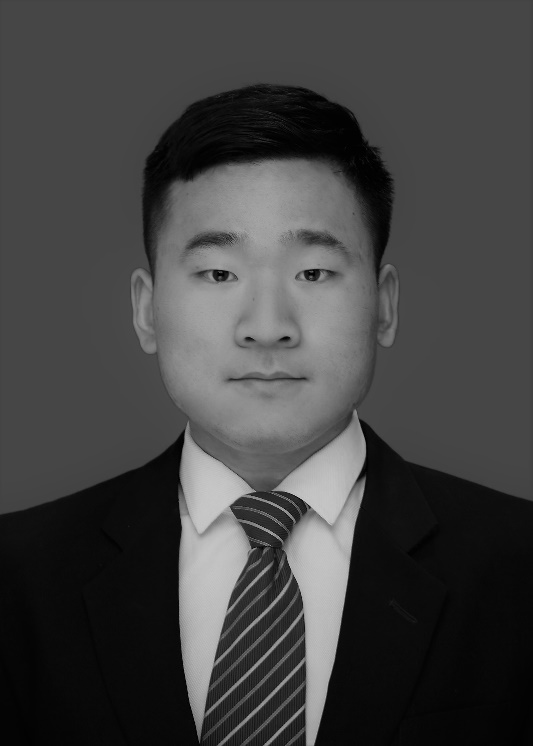}}]{Guangming Wang}
received the B.S. degree from Department of Automation from Central South University, Changsha, China, in 2018. He is currently pursuing the Ph.D. degree in Control Science and Engineering with Shanghai Jiao Tong University. His current research interests include SLAM and computer vision, in particular, visual odometry and 3D LiDAR odometry.
\end{IEEEbiography}
\vspace{-10mm}
\begin{IEEEbiography}[{\includegraphics[width=0.9in,height=1.3in,clip]{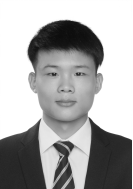}}]{Zhiheng Feng}
is currently pursuing the B.S. degree in Department of Automation, Shanghai Jiao Tong University. His latest research interests include SLAM and computer vision. 
\end{IEEEbiography}
\vspace{-10mm}
\begin{IEEEbiography}[{\includegraphics[width=0.9in,height=1.3in,clip]{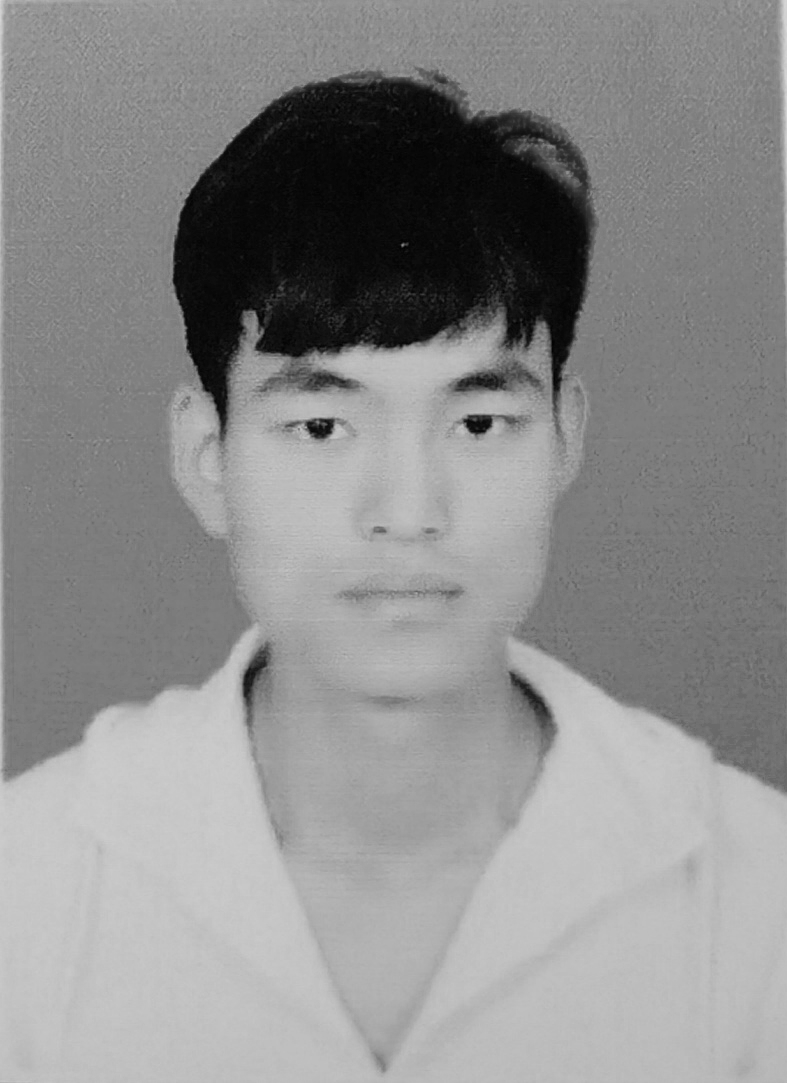}}]{Chaokang Jiang}
received the B.S. degree from the Department of Engineering, Jingdezhen Ceramic Institute College of Technology and Art, Jingdezhen, China, in 2020. He is currently pursuing the M.E. degree in control science and engineering with China University of Mining and Technology. His current research interests include SLAM and computer vision.
\end{IEEEbiography}
 \vspace{-10mm}
\begin{IEEEbiography}[{\includegraphics[width=0.9in,height=1.3in,clip]{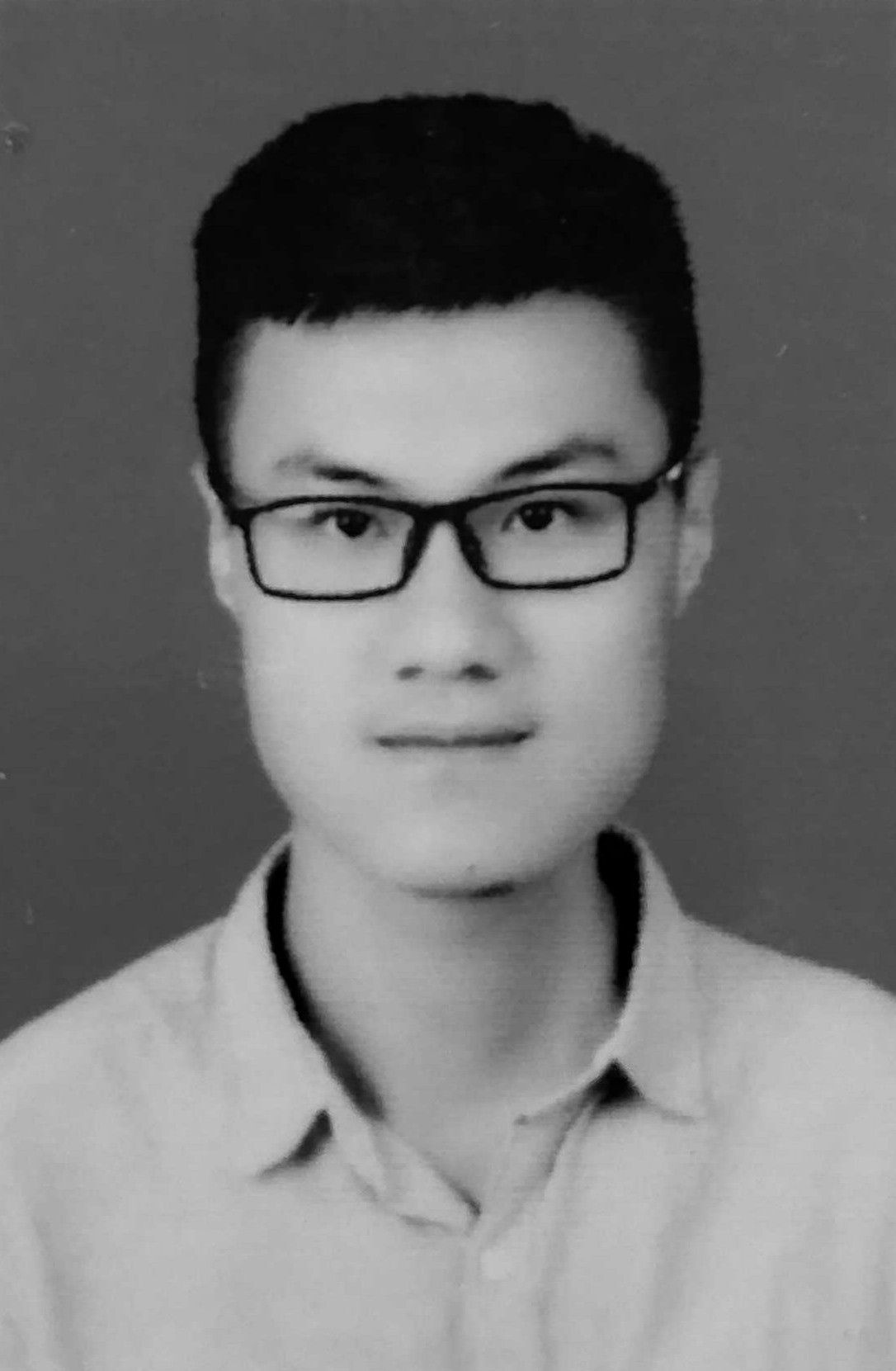}}]{Xinrui Wu}
received the B.S. degree from the
Department of Automation, Shanghai Jiao Tong University, Shanghai, China, in 2021, where he is currently pursuing the M.S. degree in Control
Science and Engineering. His latest research
interests include SLAM and computer vision. 
\end{IEEEbiography}
 \vspace{-10mm}
\begin{IEEEbiography}[{\includegraphics[width=1in,height=1.25in,clip]{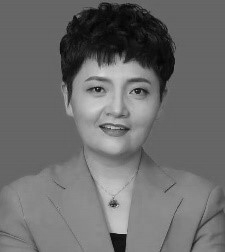}}]{Yanzi Miao} professor at the School of Information and control Engineering, China University of Mining and Technology. Her current research interests include Intelligent Perception and Fusion, Machine Vision and Active Olfaction. She received the Ph.D. degree in control science and engineering in 2009 from the China University of Mining and Technology, Xuzhou, China. As being a joint-PhD candidate and a visiting scholar, she worked in the Department of Informatics, University of Hamburg, Germany, in 2007 and 2017, respectively. She has served as the Technical Co-Chair of IEEE RCAR2019. 
\end{IEEEbiography}
 \vspace{-10mm}
\begin{IEEEbiography}[{\includegraphics[width=1in,height=1.25in,clip,keepaspectratio]{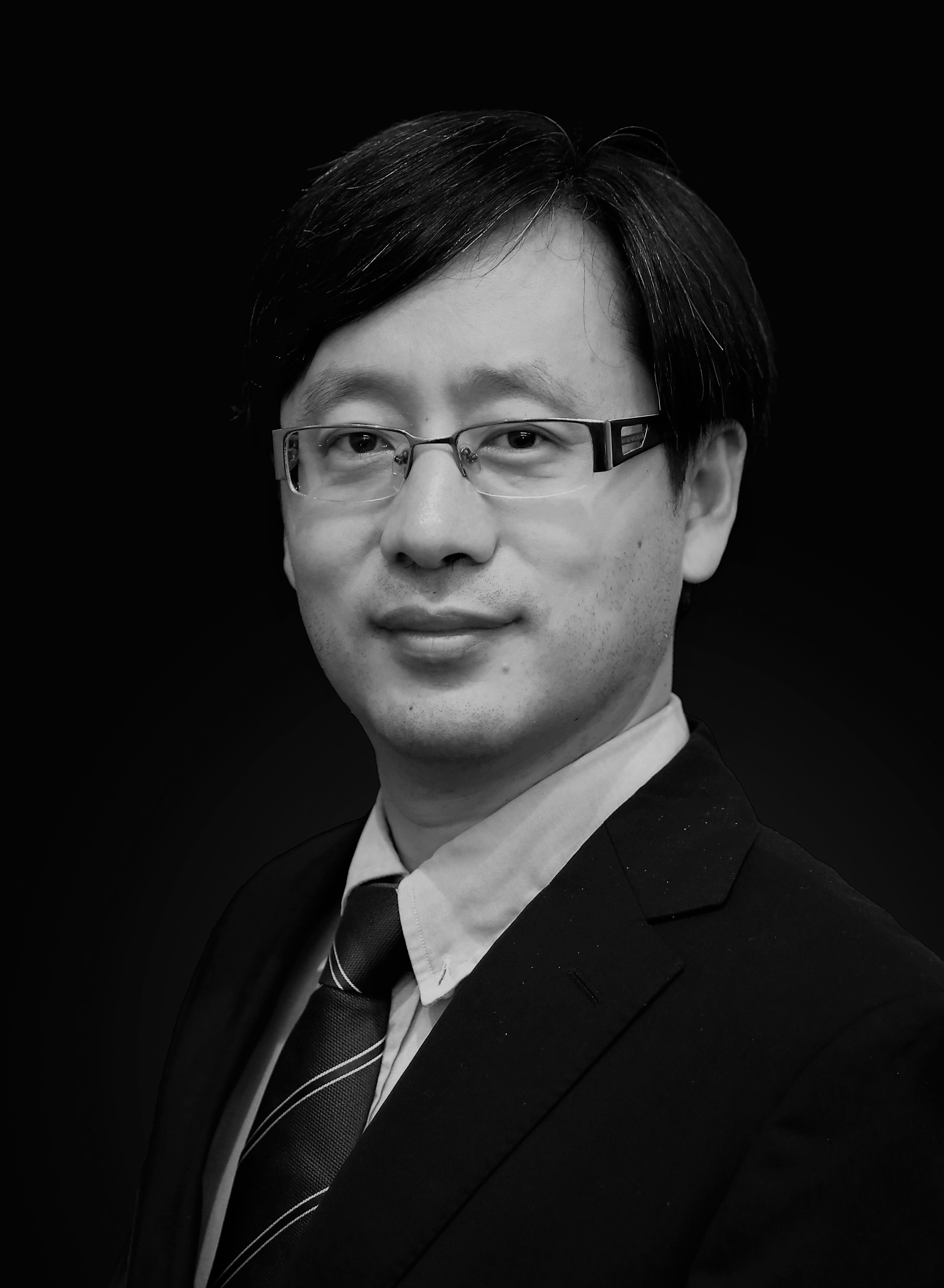}}]{Hesheng Wang} (SM’15) received the B.Eng. degree in electrical engineering from the Harbin Institute of Technology, Harbin, China, in 2002, and the M.Phil. and Ph.D. degrees in automation and computer-aided engineering from The Chinese University of Hong Kong, Hong Kong, in 2004 and 2007, respectively. He is currently a Professor with the Department of Automation, Shanghai Jiao Tong University, Shanghai, China. His current research interests include visual servoing, service robot, computer vision, and autonomous driving. 
Dr. Wang is an Associate Editor of IEEE Transactions on Automation Science and Engineering, IEEE Robotics and Automation Letters, Assembly Automation and the International Journal of Humanoid Robotics, a Technical Editor of the IEEE/ASME Transactions on Mechatronics, an Editor of Conference Editorial Board (CEB) of IEEE Robotics and Automation Society. He served as an Associate Editor of the IEEE Transactions on Robotics from 2015 to 2019. He was the General Chair of IEEE ROBIO 2022 and IEEE RCAR 2016, and the Program Chair of the IEEE ROBIO 2014 and IEEE/ASME AIM 2019. He will be the General Chair of IEEE/RSJ IROS 2025.
\end{IEEEbiography}

\end{document}